\documentclass[11pt,a4paper,english]{article}
\usepackage{mathptmx}
\usepackage{courier}
\usepackage[T1]{fontenc}
\usepackage[utf8]{inputenc}
\usepackage{array}
\usepackage{bbding}
\usepackage{url}
\usepackage{multirow}
\usepackage{graphicx}
\usepackage{microtype}

\makeatletter

\pdfpageheight\paperheight
\pdfpagewidth\paperwidth

\newcommand{\noun}[1]{\textsc{#1}}
\providecommand{\tabularnewline}{\\}

\usepackage{acl}

\makeatother

\usepackage{babel}
\begin{document}
\title{Triangular Transfer: Freezing the Pivot for Triangular Machine Translation}
\author{Meng Zhang, Liangyou Li, Qun Liu\\
Huawei Noah's Ark Lab\\
\texttt{\{zhangmeng92, liliangyou, qun.liu\}@huawei.com}}
\maketitle
\begin{abstract}
Triangular machine translation is a special case of low-resource machine
translation where the language pair of interest has limited parallel
data, but both languages have abundant parallel data with a pivot
language. Naturally, the key to triangular machine translation is
the successful exploitation of such auxiliary data. In this work,
we propose a transfer-learning-based approach that utilizes all types
of auxiliary data. As we train auxiliary source-pivot and pivot-target
translation models, we initialize some parameters of the pivot side
with a pre-trained language model and freeze them to encourage both
translation models to work in the same pivot language space, so that
they can be smoothly transferred to the source-target translation
model. Experiments show that our approach can outperform previous
ones.
\end{abstract}

\section{Introduction}

Machine translation (MT) has achieved promising performance when large-scale
parallel data is available. Unfortunately, the abundance of parallel
data is largely limited to English, which leads to concerns on the
unfair deployment of machine translation service across languages.
In turn, researchers are increasingly interested in non-English-centric
machine translation approaches \cite{fan_beyond_2021}.

Triangular MT \cite{kim_pivot-based_2019,ji_cross-lingual_2020} has
the potential to alleviate some data scarcity conditions when the
source and target languages both have a good amount of parallel data
with a pivot language (usually English). \citet{kim_pivot-based_2019}
have shown that transfer learning is an effective approach to triangular
MT, surpassing generic approaches like multilingual MT.

However, previous works have not fully exploited all types of auxiliary
data (Table \ref{data-usage}). For example, it is reasonable to assume
that the source, target, and pivot language all have much monolingual
data because of the notable size of parallel data between source-pivot
and pivot-target.

\begin{table}
\begin{centering}
{\footnotesize{}}%
\begin{tabular}{|c|c|c|c|c|c|c|}
\hline 
{\footnotesize{}approach} & \texttt{\scriptsize{}X} & \texttt{\scriptsize{}Y} & \texttt{\scriptsize{}Z} & \texttt{\scriptsize{}X}{\scriptsize{}-}\texttt{\scriptsize{}Z} & \texttt{\scriptsize{}Z}{\scriptsize{}-}\texttt{\scriptsize{}Y} & \texttt{\scriptsize{}X}{\scriptsize{}-}\texttt{\scriptsize{}Y}\tabularnewline
\hline 
\hline 
{\footnotesize{}no transfer} &  &  &  &  &  & {\scriptsize{}\Checkmark{}}\tabularnewline
\hline 
{\footnotesize{}pivot translation} &  &  &  & {\scriptsize{}\Checkmark{}} & {\scriptsize{}\Checkmark{}} & \tabularnewline
\hline 
{\footnotesize{}step-wise pre-training} &  &  &  & {\scriptsize{}\Checkmark{}} & {\scriptsize{}\Checkmark{}} & {\scriptsize{}\Checkmark{}}\tabularnewline
\hline 
{\footnotesize{}shared target transfer} & {\scriptsize{}\Checkmark{}} &  & {\scriptsize{}\Checkmark{}} &  & {\scriptsize{}\Checkmark{}} & {\scriptsize{}\Checkmark{}}\tabularnewline
\hline 
{\footnotesize{}shared source transfer} &  & {\scriptsize{}\Checkmark{}} & {\scriptsize{}\Checkmark{}} & {\scriptsize{}\Checkmark{}} &  & {\scriptsize{}\Checkmark{}}\tabularnewline
\hline 
{\footnotesize{}simple triang. transfer} &  &  & {\scriptsize{}\Checkmark{}} & {\scriptsize{}\Checkmark{}} & {\scriptsize{}\Checkmark{}} & {\scriptsize{}\Checkmark{}}\tabularnewline
\hline 
{\footnotesize{}triangular transfer} & {\scriptsize{}\Checkmark{}} & {\scriptsize{}\Checkmark{}} & {\scriptsize{}\Checkmark{}} & {\scriptsize{}\Checkmark{}} & {\scriptsize{}\Checkmark{}} & {\scriptsize{}\Checkmark{}}\tabularnewline
\hline 
\end{tabular}{\footnotesize\par}
\par\end{centering}
\caption{Data usage of different approaches (Section \ref{subsec:Baselines}).
\texttt{X}, \texttt{Y}, and \texttt{Z} represent source, target, and
pivot language, respectively. Our triangular transfer uses all types
of data.}

\label{data-usage}
\end{table}

In this work, we propose a transfer-learning-based approach that exploits
all types of auxiliary data. During the training of auxiliary models
on auxiliary data, we design parameter freezing mechanisms that encourage
the models to compute the representations in the same pivot language
space, so that combining parts of auxiliary models gives a reasonable
starting point for finetuning on the source-target data. We verify
the effectiveness of our approach with a series of experiments.

\section{Approach}

We first present a preliminary approach that is a simple implementation
of our basic idea, for ease of understanding. We then present an enhanced
version that achieves better performance. For notation purpose, we
use \texttt{X}, \texttt{Y}, and \texttt{Z} to represent source, target,
and pivot language, respectively.

\begin{figure}
\begin{centering}
\includegraphics{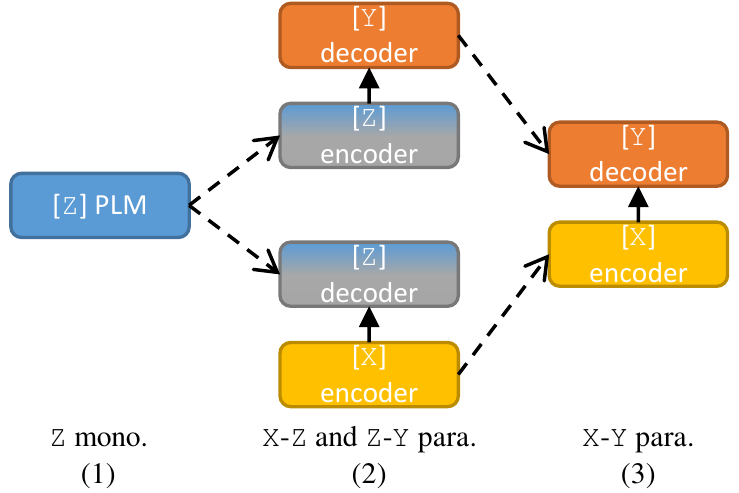}
\par\end{centering}
\caption{Simple triangular transfer. Dashed lines represent parameter initialization.
The gray color within some blocks indicates some parameters are frozen
according to the freezing strategy (Section \ref{subsec:Freezing-Strategy}).
Other colors represent trainable parameters in different languages.
Below the diagram shows the data used in each step.}

\label{preliminary}
\end{figure}

\subsection{Simple Triangular Transfer}

We show the illustration of the preliminary approach in Figure \ref{preliminary},
called simple triangular transfer. In Step (1), we prepare a pre-trained
language model (PLM) with the pivot language monolingual data. We
consider this PLM to define a representation space for the pivot language,
and we would like subsequent models to stick to this representation
space. In order to achieve this, we freeze certain parameters in Step
(2) as we train source-pivot and pivot-target translation models,
which are partly initialized by the PLM. For example, the pivot-target
translation model has the pivot language on the source side, so the
encoder is initialized by the PLM, and some (or all) of its parameters
are frozen. This ensures that the encoder produces representations
in the pivot language space, and the decoder has to perform translation
in this space. Likewise, the encoder in the source-pivot translation
model needs to learn to produce representations in the same space.
Therefore, when the pivot-target decoder combines with the source-pivot
encoder in Step (3), they could cooperate more easily in the space
defined in Step (1).

We experimented with RoBERTa \cite{liu_roberta_2019} and BART \cite{lewis_bart_2020}
as the PLMs. We found that simple triangular transfer attains about
0.8 higher BLEU by using BART instead of RoBERTa. In contrast, we
found that dual transfer \cite{zhang_two_2021}, one of our baselines,
performs similarly with BART and RoBERTa. When used to initialize
decoder parameters, RoBERTa has to leave the cross attention parameters
randomly initialized, which may explain the superiority of BART for
our approach, while dual transfer does not involve initializing decoder
parameters. Therefore, we choose BART as our default PLM.

\begin{figure*}
\begin{centering}
\includegraphics{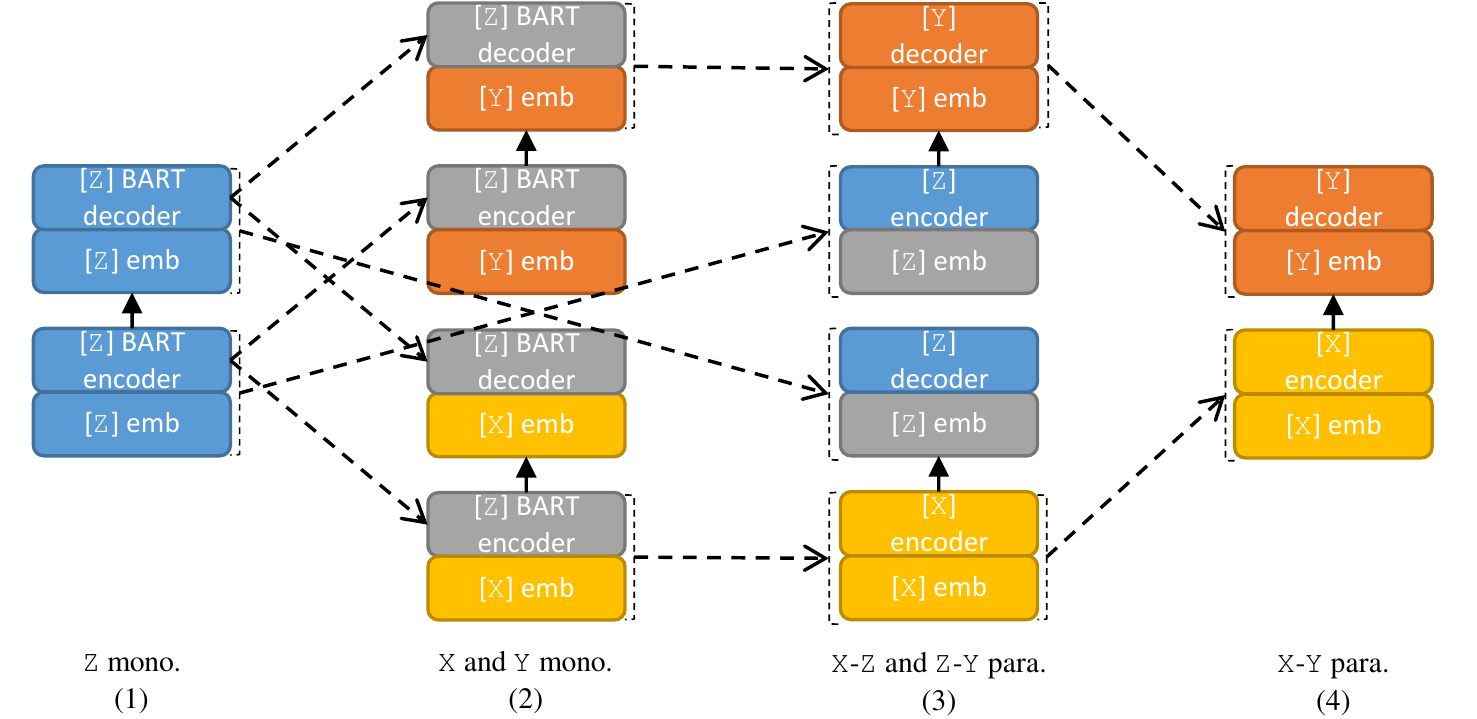}
\par\end{centering}
\caption{Triangular transfer. Dashed lines represent parameter initialization.
The gray color indicates the parameters are frozen. In Step (3) the
gray color shows one of the possible freezing strategies (Section
\ref{subsec:Freezing-Strategy}).}

\label{enhanced}
\end{figure*}

\subsection{Triangular Transfer}

A limitation of simple triangular transfer is that it does not utilize
monolingual data of the source and target languages. A naive way is
to prepare source and target PLMs and use them to initialize source-pivot
encoder and pivot-target decoder, respectively. However, this leads
to marginal improvement for the final source-target translation performance
(Section \ref{subsec:Using-Monolingual-Data}). This is likely because
the source, target, and pivot PLMs are trained independently, so their
representation spaces are isolated.

Therefore, we intend to train source and target PLMs in the pivot
language space as well. To this end, we design another initialization
and freezing step inspired by \citet{zhang_two_2021}, as shown in
Figure \ref{enhanced}. In this illustration, we use BART as the PLM.
Step (2) is the added step of preparing BART models in the source
and target languages. As the BART body parameters are inherited from
the pivot language BART and frozen, the source and target language
BART embeddings are trained to lie in the pivot language space. Then
in Step (3), every part of the translation models can be initialized
in the pivot language space. Again, we freeze parameters in the pivot
language side to ensure the representations do not drift too much.

\subsection{Freezing Strategy\label{subsec:Freezing-Strategy}}

There are various choices when we freeze parameters in the pivot language
side of the source-pivot and pivot-target translation models. Take
the encoder of the pivot-target translation model as the example.
In one extreme, we can freeze the embeddings only; this is good for
the optimization of pivot-target translation, but may result in a
space that is far away from the pivot language space given by the
pivot PLM. In the other extreme, we can freeze the entire encoder,
which clearly hurts the pivot-target translation performance. This
is hence a trade-off. We experiment with multiple freezing strategies
between the two extremes, i.e., freezing a given number of layers.
We always ensure that the number of frozen layers is the same for
the decoder of the source-pivot translation model.

Besides layer-wise freezing, we also try component-wise freezing inspired
by \citet{li_multilingual_2021}. In their study, they found that
some components like layer normalization and decoder cross attention
are necessary to finetune, while others can be frozen. In particular,
we experiment with three strategies based on their findings of the
most effective ones in their task. These strategies apply to Step
(3) of triangular transfer.

\paragraph*{LNA-E,D}

All layer normalization, encoder self attention, decoder cross attention
can be finetuned. Others are frozen.

\paragraph*{LNA-D}

All encoder parameters, decoder layer normalization and cross attention
can be finetuned.

\paragraph*{LNA-e,D}

Use LNA-D when training the source-pivot model. When training the
pivot-target model, freeze encoder embeddings in addition to LNA-D.

\section{Experiments}

\begin{table}
\begin{centering}
\begin{tabular}{|c|c|}
\hline 
language code & \# sentence (pair)\tabularnewline
\hline 
\hline 
En-De & 3.1m\tabularnewline
\hline 
Fr-En & 29.5m\tabularnewline
\hline 
Fr-De & 247k\tabularnewline
\hline 
Zh-En & 11.9m\tabularnewline
\hline 
Zh-De & 189k\tabularnewline
\hline 
En & 93.9m\tabularnewline
\hline 
De & 100.0m\tabularnewline
\hline 
Fr & 44.6m\tabularnewline
\hline 
Zh & 20.0m\tabularnewline
\hline 
\end{tabular}
\par\end{centering}
\caption{Training data statistics.}

\label{data}
\end{table}

\subsection{Setup}

We conduct experiments on French (Fr) $\rightarrow$ German (De) and
Chinese (Zh) $\rightarrow$ German (De) translation, with English
(En) as the pivot language. Training data statistics is shown in Table
\ref{data}. The evaluation metric is computed by SacreBLEU\footnote{SacreBLEU signature: BLEU+case.mixed+numrefs.1+ smooth.exp+tok.13a+version.1.4.12.}
\cite{post_call_2018}. All approaches use Transformer base \cite{vaswani_attention_2017}
as the translation model, but note that pivot translation needs two
translation models for decoding, equivalently doubling the number
of parameters. Further details can be found in the appendix.

\subsection{Baselines\label{subsec:Baselines}}

We compare with several baselines as follows.

\paragraph*{No transfer}

This baseline directly trains on the source-target parallel data.

\paragraph*{Pivot translation}

Two-pass decoding by source-pivot and pivot-target translation.

\paragraph*{Step-wise pre-training}

This is one of the approaches in \cite{kim_pivot-based_2019}. It
is simple and robust, and has been shown to outperform multilingual
MT. It trains a source-pivot translation model and uses the encoder
to initialize the encoder of a pivot-target translation model. In
order to make this possible, these two encoders need to use a shared
source-pivot vocabulary. Then the pivot-target translation model is
trained while keeping its encoder frozen. Finally the model is finetuned
on source-target parallel data.

\paragraph*{Shared target dual transfer}

Dual transfer \cite{zhang_two_2021} is a general transfer learning
approach to low-resource machine translation. When applied to triangular
MT, it cannot utilize both source-pivot and pivot-target parallel
data. Shared target dual transfer uses pivot-target auxiliary translation
model and does not exploit source-pivot parallel data.

\paragraph*{Shared source dual transfer}

The shared source version uses source-pivot translation model for
transfer and does not exploit pivot-target parallel data.

\begin{table}
\begin{centering}
\begin{tabular}{|c|c|}
\hline 
approach & BLEU\tabularnewline
\hline 
\hline 
no transfer & 13.49\tabularnewline
\hline 
pivot translation through no transfer & 18.99\tabularnewline
\hline 
step-wise pre-training & 18.49\tabularnewline
\hline 
shared target transfer & 18.88\tabularnewline
\hline 
shared source transfer & 18.89\tabularnewline
\hline 
triangular transfer & 19.91\tabularnewline
\hline 
\end{tabular}
\par\end{centering}
\caption{Comparison with baselines on Fr$\rightarrow$De. Our triangular transfer
is significantly better ($p<0.01$) than baselines by paired bootstrap
resampling \cite{koehn_statistical_2004}.}

\label{main-results}
\end{table}

\begin{table}
\begin{centering}
\begin{tabular}{|c|c|}
\hline 
approach & BLEU\tabularnewline
\hline 
\hline 
no transfer & 11.39\tabularnewline
\hline 
pivot translation through no transfer & 12.91\tabularnewline
\hline 
triangular transfer & 16.03\tabularnewline
\hline 
\end{tabular}
\par\end{centering}
\caption{Comparison with no transfer and pivot translation on Zh$\rightarrow$De.}

\label{main-results-Zh-De}
\end{table}

\subsection{Main Results}

We present the performance of our approach and the baselines on Fr$\rightarrow$De
in Table \ref{main-results}. The no transfer baseline performs poorly
because it is trained on a small amount of parallel data. The other
baselines perform much better. Among them, pivot translation attains
the best performance in terms of BLEU, at the cost of doubled latency.
Our approach can outperform all the baselines.

Taking pivot translation as the best baseline, we further evaluate
our approach on Zh$\rightarrow$De. Results in Table \ref{main-results-Zh-De}
show that the performance improvement of our approach is larger for
this translation direction.

\subsection{The Effect of Freezing Strategies}

\begin{table}
\begin{centering}
\begin{tabular}{|c||c|c||c|}
\hline 
strategy & Fr-En & En-De & Fr-De\tabularnewline
\hline 
\hline 
$L=0$ & 31.42 & 20.95 & 19.62\tabularnewline
\hline 
$L=1$ & 31.41 & 20.98 & 19.76\tabularnewline
\hline 
$L=2$ & 31.55 & 20.56 & 19.71\tabularnewline
\hline 
$L=3$ & 31.06 & 20.54 & 19.91\tabularnewline
\hline 
$L=4$ & 30.92 & 20.22 & 19.68\tabularnewline
\hline 
$L=5$ & 30.39 & 19.95 & 19.21\tabularnewline
\hline 
$L=6$ & 30.31 & 19.11 & 19.02\tabularnewline
\hline 
\hline 
LNA-E,D & 28.72 & 17.92 & 17.97\tabularnewline
\hline 
LNA-D & 31.08 & 20.23 & 18.75\tabularnewline
\hline 
LNA-e,D & 31.08 & 19.97 & 18.25\tabularnewline
\hline 
\end{tabular}
\par\end{centering}
\caption{BLEU scores of different freezing strategies for triangular transfer.
For layer-wise freezing, the embeddings and the lowest $L$ layers
of the pivot side network are frozen. If $L=0$, only the embeddings
are frozen.}

\label{freezing-strategies}
\end{table}

From Table \ref{freezing-strategies}, we can observe the effect of
different freezing strategies. For layer-wise freezing, we see a roughly
monotonic trend of the Fr-En and En-De performance with respect to
the number of frozen layers: The more frozen layers, the lower their
BLEU scores. However, the best Fr-De performance is achieved with
$L=3$. This indicates the trade-off between the auxiliary models'
performance and the pivot space anchoring. For component-wise freezing,
the Fr-En and En-De performance follows a similar trend, but the Fr-De
performance that we ultimately care about is not as good.

\subsection{Using Monolingual Data\label{subsec:Using-Monolingual-Data}}

\begin{table}
\begin{centering}
\begin{tabular}{|c|c|}
\hline 
approach & BLEU\tabularnewline
\hline 
\hline 
pivot translation through no transfer & 18.99\tabularnewline
\hline 
pivot translation through \noun{bert}2\noun{bert} & 19.06\tabularnewline
\hline 
\hline 
shared target transfer & 18.88\tabularnewline
\hline 
shared target transfer + naive mono. & 18.93\tabularnewline
\hline 
\hline 
shared source transfer & 18.89\tabularnewline
\hline 
shared source transfer + naive mono. & 18.97\tabularnewline
\hline 
\hline 
simple triang. transfer & 18.96\tabularnewline
\hline 
simple triang. transfer + naive mono. & 19.00\tabularnewline
\hline 
triangular transfer & 19.62\tabularnewline
\hline 
\end{tabular}
\par\end{centering}
\caption{Naive ways of using auxiliary monolingual data do not bring clear
improvement. Our approaches freeze embeddings as the freezing strategy
in this table.}

\label{using-monolingual-data}
\end{table}

Table \ref{using-monolingual-data} shows the effect of different
ways of using monolingual data. The naive way is to prepare PLMs with
monolingual data and initialize the encoder or decoder where needed.
For pivot translation, this is known as \noun{bert}2\noun{bert} \cite{rothe_leveraging_2020}
for the source-pivot and pivot-target translation models. For dual
transfer, parts of the auxiliary models can be initialized by PLMs
(e.g., for shared target transfer, the pivot-target decoder is initialized).
For Step (2) in simple triangular transfer, we can also initialize
the pivot-target decoder and source-pivot encoder with PLMs. However,
none of the above methods shows clear improvement. This is likely
because these methods only help the auxiliary translation models to
train, which is not necessary as they can be trained well with abundant
parallel data already. In contrast, our design of Step (2) in triangular
transfer additionally helps the auxiliary translation models to stay
in the pivot language space.

\subsection{Pivot-Based Back-Translation}

\begin{table}
\begin{centering}
\begin{tabular}{|c|c|}
\hline 
approach & BLEU\tabularnewline
\hline 
\hline 
no transfer & 18.74\tabularnewline
\hline 
shared target transfer & 20.53\tabularnewline
\hline 
shared source transfer & 20.73\tabularnewline
\hline 
triangular transfer & 20.84\tabularnewline
\hline 
\end{tabular}
\par\end{centering}
\caption{BLEU scores from training with pivot-based back-translation.}

\label{pivot-based-BT}
\end{table}

Following \citet{kim_pivot-based_2019}, we generate synthetic parallel
Fr-De data with pivot-based back-translation \cite{bertoldi_phrase-based_2008}.
Specifically, we use a no transfer En$\rightarrow$Fr model to translate
the English side of En-De data into French, and the authentic Fr-De
data are oversampled to make the ratio of authentic and synthetic
data to be 1:2. Results in Table \ref{pivot-based-BT} show that triangular
transfer and dual transfer clearly outperform the no transfer baseline.

\section{Conclusion}

In this work, we propose a transfer-learning-based approach that utilizes
all types of auxiliary data, including both source-pivot and pivot-target
parallel data, as well as involved monolingual data. We investigate
different freezing strategies for training the auxiliary models to
improve source-target translation, and achieve better performance
than previous approaches.

\bibliographystyle{acl_natbib}
\bibliography{acl2022-zm}

\appendix

\section{Data and Preprocessing}

\begin{table*}
\begin{centering}
\begin{tabular}{|c|c|c|c|c|}
\hline 
lang. & source & train & dev & test\tabularnewline
\hline 
\hline 
\multirow{2}{*}{En-De} & \multirow{2}{*}{WMT 2019} & Europarl v9, News Commentary v14, & \multirow{2}{*}{newstest2011} & \multirow{2}{*}{newstest2012}\tabularnewline
 &  & Document-split Rapid corpus &  & \tabularnewline
\hline 
\multirow{2}{*}{Fr-En} & \multirow{2}{*}{WMT 2015} & Europarl v7, News Commentary v10, & \multirow{2}{*}{newstest2011} & \multirow{2}{*}{newstest2012}\tabularnewline
 &  & UN corpus, $10^{9}$ French-English corpus &  & \tabularnewline
\hline 
Fr-De & WMT 2019 & News Commentary v14, newstest2008-2010 & newstest2011 & newstest2012\tabularnewline
\hline 
Zh-En & ParaCrawl & ParaCrawl v9 & newsdev2017 & newstest2017\tabularnewline
\hline 
Zh-De & WMT 2021 & News Commentary v16 - dev - test & 3k split & 3k split\tabularnewline
\hline 
\end{tabular}
\par\end{centering}
\caption{Parallel data source.}

\label{parallel-data-source}
\end{table*}

\begin{table*}
\begin{centering}
\begin{tabular}{|c|c|c|}
\hline 
lang. & source & name\tabularnewline
\hline 
\hline 
En & WMT 2018 & News Crawl 2014-2017\tabularnewline
\hline 
De & WMT 2021 & 100m subset from WMT 2021\tabularnewline
\hline 
\multirow{2}{*}{Fr} & \multirow{2}{*}{WMT 2015} & Europarl v7, News Commentary v10,\tabularnewline
 &  & News Crawl 2007-2014, News Discussions\tabularnewline
\hline 
Zh & WMT 2021 & News Crawl, Zh side of parallel data\tabularnewline
\hline 
\end{tabular}
\par\end{centering}
\caption{Monolingual data source.}

\label{monolingual-data-source}
\end{table*}

We gather data from WMT and ParaCrawl, shown in Tables \ref{parallel-data-source}
and \ref{monolingual-data-source}.

We use \texttt{jieba}\footnote{\texttt{\url{https://github.com/fxsjy/jieba}}}
for Chinese word segmentation, and Moses\footnote{\url{https://github.com/moses-smt/mosesdecoder}}
scripts for punctuation normalization and tokenization of other languages.
The corpora are deduplicated. Each language is encoded with byte pair
encoding (BPE) \cite{sennrich_neural_2016} with 32k merge operations.
The BPE codes and vocabularies are learned on each language’s monolingual
data, and then used to segment parallel data. Sentences with more
than 128 subwords are removed. Parallel sentences are cleaned with
length ratio 1.5 (length counted by subwords).

\section{Hyperparameters}

Our implementation is based on \texttt{fairseq} \cite{ott_fairseq_2019}.
We share decoder input and output embeddings \cite{press_using_2017}.
The optimizer is Adam. Dropout and label smoothing are both set to
0.1. The batch size is 6,144 per GPU and we train on 8 GPUs. The peak
learning rate is $5\times10^{-4}$ for the no transfer baseline and
auxiliary models, $1\times10^{-4}$ for the Fr$\rightarrow$De model
of step-wise pre-training and dual transfer, and $7\times10^{-5}$
for the last step of triangular transfer. The learning rate warms
up for 4,000 steps, and then follows inverse square root decay. Early
stopping happens when the development BLEU does not improve for 10
epochs.

RoBERTa and BART models use exactly the same architecture as Transformer
base. The mask ratio is 15\%. The batch size is 256 sentences per
GPU, and each sentence contains up to 128 tokens. The learning rate
warms up for 10,000 steps to the peak $5\times10^{-4}$, and then
follows polynomial decay. They are trained for 125k steps.

We use beam size of 5 for decoding, including for pivot translation
and pivot-based back-translation.
\end{document}